\documentclass[runningheads]{llncs}
\usepackage[T1]{fontenc}
\usepackage{graphicx}
\usepackage{hyperref}
\usepackage{color}

\begin{document}

\title{A First Step in Using Machine Learning Methods to Enhance Interaction Analysis for Embodied Learning Environments}

\titlerunning{Using ML to Enhance IA in Embodied Learning}

\author{Joyce Fonteles\inst{1}\orcidID{0000-0001-9862-8960} 
\and Eduardo Davalos\inst{1}\orcidID{0000-0001-7190-7273} 
\and Ashwin T. S.\inst{1}\orcidID{0000-0002-1690-1626} 
\and Yike Zhang\inst{1}\orcidID{0000-0003-3503-2996} 
\and Mengxi Zhou\inst{2}\orcidID{0009-0003-6902-0325} 
\and Efrat Ayalon\inst{1}\orcidID{0009-0006-6679-2452} 
\and Alicia Lane\inst{1}\orcidID{0009-0002-1589-621X} 
\and Selena Steinberg\inst{2}\orcidID{0000-0003-0032-8957}
\and Gabriella Anton\inst{1}\orcidID{0000-0002-0117-053} 
\and Joshua Danish\inst{2}\orcidID{0000-0001-5119-5897} 
\and Noel Enyedy\inst{1}\orcidID{0000-0001-7662-5654} 
\and Gautam Biswas\inst{1}\orcidID{0000-0002-2752-3878} }

\authorrunning{Fonteles et al.}

\institute{Vanderbilt University, Nashville TN 37240, USA \\
\email{\{joyce.h.fonteles,gautam.biswas\}@Vanderbilt.edu}\\
\and Indiana University, Bloomington IN 47405, USA
}

\maketitle

\begin{abstract}
Investigating children's embodied learning in mixed-reality environments, where they collaboratively simulate scientific processes, requires analyzing complex multimodal data to interpret their learning and coordination behaviors. Learning scientists have developed Interaction Analysis (IA) methodologies for analyzing such data, but this requires researchers to watch hours of videos to extract and interpret students’ learning patterns. Our study aims to simplify researchers' tasks, using Machine Learning and Multimodal Learning Analytics to support the IA processes. Our study combines machine learning algorithms and multimodal analyses to support and streamline researcher efforts in developing a comprehensive understanding of students' scientific engagement through their movements, gaze, and affective responses in a simulated scenario. To facilitate an effective researcher-AI partnership, we present an initial case study to determine the feasibility of visually representing students' states, actions, gaze, affect, and movement on a timeline. Our case study focuses on a specific science scenario where students learn about photosynthesis. The timeline allows us to investigate the alignment of critical learning moments identified by multimodal and interaction analysis, and uncover insights into students' temporal learning progressions. 

\end{abstract}

\section{Introduction}

Embodied learning aligns with the natural ways in which humans perceive, interact, and learn from the world around them. By engaging the body in the learning process, we create richer, more immersive educational experiences where our actions, movements, and interactions contribute significantly to how we understand and internalize concepts \cite{danish2020learning}. It allows students to actively explore and embody knowledge through perception, awareness, and exploration of their environment. Embodiments not only enhance retention and a deeper understanding of abstract or complex concepts; it leverages the power of immersive experiences to make education more engaging and impactful \cite{enyedy2014learning}. 

Embodied learning data analysis presents a great challenge due to the complexity of monitoring student groups spatially and temporally. Conventional educational settings focus mostly on verbal communication and digital system interactions. Meanwhile, embodied learning necessitates the capture of non-verbal cues and body movements in 3D space, along with conversations and simulation logs \cite{danish2020learning}.  Interaction Analysis (IA) is one of the main approaches employed by learning scientists because it can unravel deep insights and nuanced interactions captured in video data \cite{hall2015interaction}. IA yields valuable insights, but its manual processes are time-consuming and demand substantial human resources. Therefore, recent advances in Machine Learning (ML) and Multimodal Learning Analytics (MMLA) make it easier to leverage algorithms to support human analysis, with the idea that the combination will allow researchers and educators to gain a nuanced understanding of how learners engage with content, facilitating feedback, assessment, and an enriched comprehension of the learning process \cite{martinez2018physical}.

Observing and tracking embodied learning scenarios generates substantial volumes of multimodal data, which is a reflection of the diverse sensory inputs needed to capture movement, gestures, gaze, interactions, and coordination in the context of learning and problem-solving tasks \cite{andrade2017understanding}. Motion tracking data, detailing the positions and orientations of body parts, enables granular analysis of physical interactions. Gaze tracking data reveals where learners direct their attention, shedding light on points of interest or challenges. Affect detection data adds a nuanced layer by gauging learners' emotional states. Complementing these, system logs record interactions with the learning platform, simulations, or virtual environments, offering timestamps and details of actions taken. Managing all of the heterogeneous multimodal data efficiently is a complex task, demanding sophisticated computational analysis. The challenge for educational methods using AI and ML lies in addressing the complexities of multimodal data collection, alignment, and analysis to derive meaningful insights into students' individual and collaborative behaviors in a timely manner. 

Human researchers, familiar with the varied contexts of embodied learning data, are crucial for its interpretation. Unlike technology-centric approaches, we advocate for "\textit{AI-in-the-loop}" methods, emphasizing the pivotal role of humans in the analysis and interpretation process. This study makes two primary contributions. First, it applies IA to determine the most effective modalities, analyses, and visualizations for employing AI to aid human interpretation of student behavior. Second, it introduces an interactive visual timeline that displays MMLA results, tailored to augment IA. This timeline represents students' movements, necessitating data processing from multiple cameras for accurate student re-identification and face tracking. Our findings reveal that these environments provoke emotional responses distinct from those observed in traditional computer-based learning settings. Additionally, we propose an innovative approach for discretizing gaze toward moving objects in 3D spaces, significantly contributing to the field of Artificial Intelligence in Education.
\section{Background}
\subsection{Embodied Learning and Interaction Analysis}
Embodied learning activities leverage the movement of students' bodies in the teaching and learning of conceptual and disciplinary ideas. These activities are grounded in the assumption that parallels can be drawn between bodily experiences and conceptual learning \cite{kersting2021growing}. 
In science education, many embodied activity designs rely on educational technologies like virtual reality (VR) and mixed reality (MR) to immerse students in a particular scientific system, enabling feelings of presence. 
The design of embodied activities can be approached from a cognitive lens, which focuses on how individual students' movements or gestures can map onto underlying conceptual ideas 
or from a sociocultural lens, which considers the ways that youth interact with each other as socially situated. In this work, we take a sociocultural approach to designing these kinds of embodied learning activities \cite{danish2020learning}. This means that we are most interested in embodiment that happens between multiple learners in a social and cultural context; as youth engage with each other, they develop meaning together through their embodiment. 

As youth participate in collaborative MR embodied activities, they must attend to and coordinate many modes, including gaze, movement, and speech \cite{steinberg2023making}. IA empirically investigates human interactions with each other and objects in contextual settings. Learning scientists often use this analytic method to make sense of collaborative, embodied learning environments, where multiple students move together in a classroom or educational setting. IA's development is theoretically grounded in several methodologies, notably conversation analysis and ethnography, and has become popular with the proliferation of audiovisual recording technologies. The capability to capture learning activities from multifaceted views/positions and iterative playback of recordings is crucial to interaction analysis, as it allows close interrogation, which is the essence of IA.
Thus, the goal of interaction analysis is to look for empirical evidence of the learning and learning process by discerning patterns/ regularities in how participants utilize resources within their natural environments and interact with each other.

\subsection{Affect and Learning}
In academics, the detection of learning-centric emotions $-$ confusion, boredom, frustration, engagement, and delight $-$ is crucial for comprehending learner behaviors and performance 
\cite{pekrun2012academic}.
While state-of-the-art computer vision algorithms can successfully identify basic and learning-centered emotions, their application in embodied learning environments presents unique challenges \cite{ts2020automatic}. In such settings, multiple students are often captured in a single video frame. Further, students are moving frequently, and this necessitates advanced techniques like re-identification for accurate emotion tracking \cite{tang2017multiple}. 

A significant shortcoming in current emotion recognition datasets is their focus on undergraduate students, with little data representing children's facial expressions, which makes it hard to detect their emotions accurately \cite{ashwin2020affective}. 
Models like High-Speed Emotion Recognition, which are trained on diverse age groups, including children, offer an alternative by quantifying emotions on a continuous scale of valence and arousal \cite{savchenko2022classifying}. These models utilize frameworks such as Russell's circumplex model and D'Mello's dynamics of emotion to translate continuous emotional states into discrete categories \cite{russell1980circumplex,d2012dynamics}. In dynamic environments, where students may walk, jump, and exhibit rapid movements, top-tier models like multitask cascaded convolutional networks (MTCNN) are preferred for face identification \cite{zhang2016joint}. However, they typically lack training for rapid movements and partial occlusions, characteristic of embodied learning. Consequently, retraining existing models or deriving new ones tailored for these complex settings is imperative for the accurate identification and analysis of student emotions.

\subsection{Gaze Detection and Interactions}
Gaze analysis has been a cornerstone in understanding how learners engage and process information, revealing insights into cognitive functions and social interactions that are critical for collaborative learning and problem-solving \cite{Vatral2022Using}. Traditionally, eye-tracking required specific hardware and controlled environments, limiting its application in actual classroom settings and affecting the authenticity of observed learning behaviors. To overcome these constraints, advances in eye-tracking technology have introduced more versatile tools, such as lightweight eye-tracking glasses like the Tobii Glasses 3. These innovations allow for observation in more natural settings, although they face challenges like limited scalability and adaptability for children. Computer-vision methods, such as L2CS-Net \cite{Abdelrahman2022L2CSNetF} and Gaze360 \cite{gaze360_2019}, offer solutions that are more suitable for the dynamic nature of classrooms, even though they may compromise some on the precision of gaze data. Despite this, the trade-off is considered acceptable for educational research, where the focus is on broader data interpretation rather than pinpoint accuracy. However, applying these methods to children remains problematic because they have not been trained on their data.

In eye-tracking research, encoding gaze data into objects of interest (OOI) helps translate raw gaze points into meaningful insights by associating them with elements in the learning environment, such as teaching aids or interactive tools. The distinction between static and dynamic OOIs presents a significant challenge, requiring sophisticated tracking and analysis techniques \cite{Davalos2023Identifying}. 3D reconstruction emerges as a promising approach to address this, enabling detailed spatial analysis of gaze patterns\cite{Li2020Visualization}. However, the task is complex, especially when relying on monocular video capture that lacks depth information, posing hurdles for accurately mapping gaze in three-dimensional spaces.

\subsection{Multimodal Visualizations}

The visualization of multimodal data, which combines various modalities such as facial expressions, interactions, and contextual information, is an active area of research in education and learning analytics. Ez-zaouia's Emodash \cite{EzZaouia2020Emodash} contributes to this field by presenting a dashboard that visualizes learners' emotions inferred from facial expressions, alongside their system interactions during online learning sessions, reinforcing how one of the main challenges in the field is finding the appropriate level of detail and timescales for visualizations. This work builds upon previous research that explored visualizing learners' performances and behaviors using primarily systems logs dashboards, as well as the design of multimodal and contextual emotional dashboards for tutors \cite{Schwendimann2017Dashboards}. 

Our design solution is distinct from the current literature. The timeline structure has been shaped by IA sessions conducted by researchers to identify key modalities and analytical approaches for interpreting student actions in embodied learning environments. Contrasting with prior research focused on computer-based learning environments, our study explores the unique dynamics of embodied learning within a MR context, where students' physical movements facilitate interaction. While our approach aligns with Ez-zaouia et al. in integrating system interactions and affect into an interactive timeline, it further examines the specific emotional responses elicited by the gamified aspects of embodied learning. We also incorporate gaze data to study students' shifting attention during activities. Our innovative presentation of multimodal data on a dynamic timeline, synchronized with video playback, is designed to enhance IA by providing AI-generated insights and interpretations to researchers.
\section{Methods}
\subsection{Study Design}

This study is part of a larger project entitled Generalized Embodied Modeling to support Science through Technology Enhanced Play (GEM-STEP). In this project, the motion-tracking technologies and mixed-reality environments display participants’ movements on a projector screen, embodying complex scientific phenomena as researchers investigate individual and collaborative learning processes \cite{danish2022designing}. The GEM-STEP research team, including two of the authors, collaborated with a fourth-grade science teacher to co-design and co-facilitate a 20-day curriculum focused on food webs and photosynthesis. The participants in the following analysis consisted of two facilitators (the teacher and one researcher) and seven consented and assented students (four boys and three girls) from diverse racial and linguistic backgrounds. The students included multilingual learners, and their home languages included English, Kurdish, and Spanish.

Considering the diverse learners in our site, technology-mediated and embodied learning environments can expand access to science content where conventional text- or discourse-based learning may not. These environments expand students’ sense-making resources, e.g., their bodies and emotions, often restricted within science learning contexts. This multimodal approach promotes students’ agency and engagement  while lowering linguistic barriers\cite{lane2024embodied}. The curriculum and the models were designed based on these approaches while aligning with local science standards. In this work we focus on the photosynthesis model, a closed-loop system with a simulation screen which alternates from day to night. It features a mouse and a tomato plant with zoomed-in chloroplasts and roots (areas that cause molecule transformations). Students must move among these locations to model interactions between the molecules they are embodying (oxygen, water, sugar and carbon dioxide) and features on the screen. 
To streamline data collection we employed a distributed streaming framework called ChimeraPy\cite{Davalos2023chimerapy}, which supports rapid deployment in the classroom and provides time-alignment across multiple data modalities. We collected video from four cameras, multiple wireless microphones, screen recordings, and system logs.

\subsection{Interaction Analysis by Researchers}

Four authors from the learning sciences completed the IA of the videos. While all four analysts were knowledgeable about embodiment and participated in the design of the GEM-STEP project, two were present at data collection and two were not, and thus were less familiar with the context and the data. We split the videos among researchers such that two authors reviewed day 1 and two authors reviewed day 2, paired such that one researcher had familiarity with the site and one did not. In this way, we hoped that we might elicit multiple, diverse perspectives on the videos. For analysis, we selected three focal students because, on the first pass, they seemed to approach the activity in diverse ways. We focused on how each student seemed to be moving and when they seemed to understand the photosynthesis content. 

During these co-watching sessions, researchers paused video-playing when observing notable events, such as instances of students’ laughter or moments when a student facilitated the modeling of others. They then discussed the significance of these moments, including whether students demonstrated evidence of learning, what evidence was, and how the learning was mediated. Additionally, they discussed moments when IA supported understanding of the learning process were highlighted, including moments of contextual interactions. Moments when affect and gaze analysis could be important were also identified. For example, students displayed different emotions: one student (pseudonym Rose) was excited when the photosynthesis process was successful, while another student (pseudonym Taylor Swift) remained very calm at these moments. One student (pseudonym DaPaw) shifted his gaze and body away from the simulation screen while successfully modeling. Human analysts marked those moments as interesting to investigate how the AI findings might align or not with IA.

\subsection{Design of Visual Timeline}
By studying the IA methods applied to the embodied learning videos, it was apparent that a contextual AI-based analysis would require documenting the system's evolving state in MR, and developing algorithms for tracking and interpreting students' actions and engagement in the context of the scientific process being enacted. To visualize all of this information, we designed a visual timeline to strategically incorporate the multimodal data and analysis methods aligned with these requirements. It integrates data from system logs that tracked students' movements, including avatar shifts and interactions with objects of interest in the MR scene. Another IA-driven insight was the need to understand how system dynamics and variables, such as day-night transitions, impacted students' actions. Analyzing students' engagement and focus informed the inclusion of gaze, offering insights into the impact of shifts between the display screen, teacher, and peers on subsequent student actions. Moreover, the timeline also used facial data derived from the video analysis to capture and document students' emotions, acknowledging their significance for learning.

A powerful visualization driven by AI and ML algorithms can be a gateway to recognizing and interpreting students' individual and collective \textit{aha} moments, signaling their insights and discoveries, which in turn can be interpreted in terms of their learning the science content. The results of our analysis had to be displayed in a way that merged the modalities into one visualization that should be clear and compelling to the human researchers, while avoiding complexities and clutter that could become tiresome. Our data visualization was initially inspired by Clara Peni'n and Jaime Serra's work from La Vanguardia called "Apoteosis 'Waka Waka'"\cite{Errea2017visual}, which visualizes different aspects of a concert on a timeline including lighting cues, visual effects, costumes, and lyrics; and extends Hervé's multimedia player\cite{Herve2015amaliajs}. 

\subsection{Analysis of System Logs}

The analysis of system logs served as a primary component in establishing the nuances of student interactions and their evolution within the MR environment. These logs serve as a temporal record, tracing the sequence of avatar changes and movements made by students throughout the learning activity. By analyzing them, it is possible to discern patterns and trends in how students navigate the virtual space, interact with different elements, and transition between avatars. Understanding the spatial and temporal aspects of students' movements within the learning environment played a crucial role in gauging their existing knowledge and ongoing comprehension processes. Within this dataset, we extracted timestamped information to capture three key dimensions of data to assist in understanding students' learning processes: \textit{(1) Students' States}: Understanding which molecules students were embodying at any given time allowed us to explore their evolving comprehension of scientific concepts much like the IA researchers did; \textit{(2) Students' Actions}: Analyzing actions intertwined with their embodiment of molecules was fundamental in gauging their understanding of the photosynthesis process; \textit{(3) System State}: Capturing the influencing variable of whether the simulation was in daylight or at night and tracking students' responses to this allowed us to discern not only if but also when they grasped the concept that photosynthesis requires sunlight. 

\subsection{Affect Detection}

\begin{figure}
\centering
\includegraphics[width=0.85\textwidth]{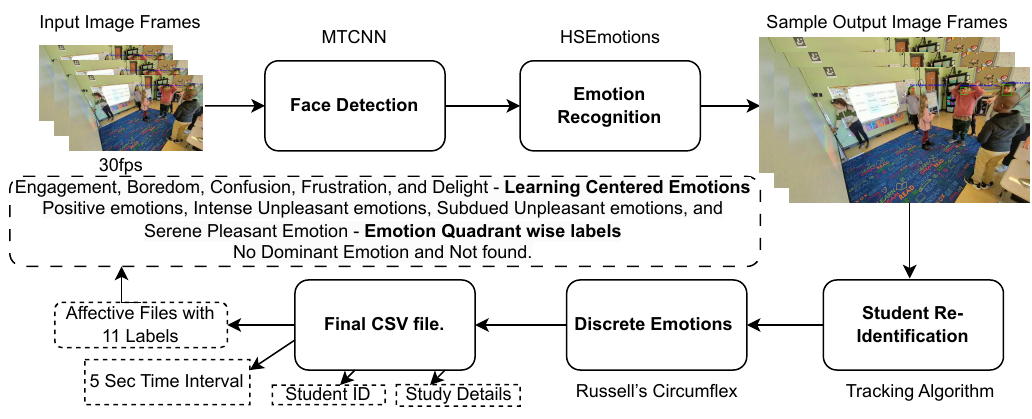}
\caption{Overview of Affect Detection Process} \label{fig1}
\end{figure}

In this study, we analyzed students' facial expressions in the embodied learning environment for emotion recognition. Figure \ref{fig1} illustrates the process, which involves face detection, followed by predicting continuous emotion scores on a valence-arousal scale. These scores are then categorized into learning-centered emotions based on Russell's circumflex of emotions \cite{russell1980circumplex} and D'Mello's dynamics of affective states \cite{d2012dynamics}. Positive emotions are assigned to the first quadrant, intense unpleasant emotions to the second quadrant, subdued unpleasant emotions to the third quadrant, and serene pleasant emotions to the fourth quadrant.

The system was initially configured to read input from a video file, initializing tools for face detection, facial landmark extraction, and emotion recognition. Notably, we fine-tuned MTCNN \cite{zhang2016joint} with thresholds of 0.8 for P-Net, R-Net, and O-Net. Additionally, we employed Dlib's facial landmark detector to precisely identify critical features on the face.

During the detailed frame-by-frame processing phase, each video frame was converted from BGR to RGB color space. MTCNN scrutinized the frame for faces, and each detected face underwent further analysis using dlib's detector for facial landmarks. The HSEmotionRecognizer processed the face region (enet\_b0\_8\_best\_afew.pt) and predicted the valance arousal values. Detailed information about each face was recorded for every frame, forming a comprehensive dataset for subsequent analysis. The system augmented the video with annotations, marking faces with bounding boxes and indicating valence and arousal scores, along with facial landmarks. This enriched video showcased the emotional analysis visually. Concurrently, it compiled this data into a structured CSV file, including the bounding boxes, valence, and arousal values. This file provided a frame-by-frame record of the emotional metrics, supporting thorough analysis and verification of the system's accuracy.

\noindent
\textbf{Student Re-Identification.} We enhanced MTCNN and HSEmotion by integrating a tracking algorithm that utilized CSV data containing frame numbers, valence, arousal, and bounding box coordinates to re-identify students across video frames. Operating at 30 fps, the algorithm maintained spatial continuity, computed bounding box centers, assigned unique IDs, and predicted positions using Euclidean distances within a set threshold. A memory component improved accuracy by compensating for minor face displacements and maintained tracking even with occlusions or movements. This approach achieved a 91\% re-identification success rate, with manual adjustments addressing the remainder.

For gaze tracking and emotion categorization, the CSV data post reidentification enabled the transformation of continuous emotion metrics into discrete states, processed in 5-second intervals to align with the frame rate. Emotions sustained over 150 frames were deemed significant, except for delight, which required 60 frames. Minimal facial expressions were labeled as ``Engaged Concentration," consistent with educational emotion research \cite{d2012dynamics}. Emotions were classified into Engagement, Boredom, Confusion, Frustration, and Delight, or by valence-arousal quadrants, with 11 labels in total, including `NotFound' and `NoDominantEmotion' for cases where faces were not visible (see Figure \ref{fig1}). 

\subsection{Gaze Estimation}

\begin{figure}
    \centering
    \includegraphics[width=0.85\textwidth]{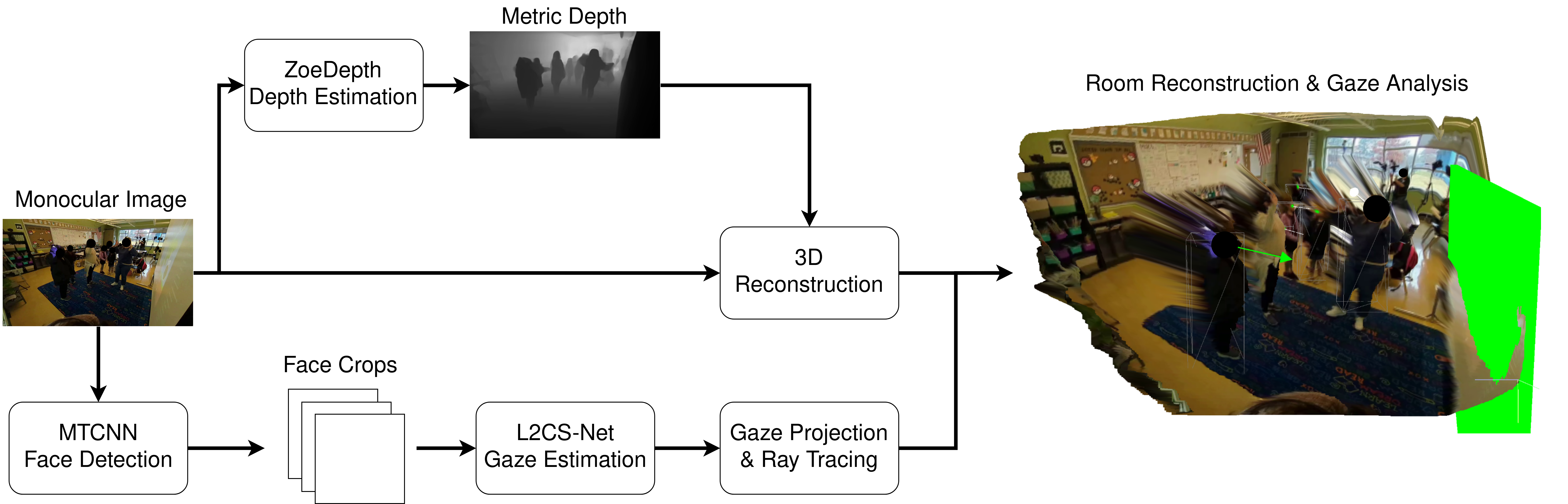}
    \caption{Gaze Estimation Pipeline.}
    \label{fig:gaze_estimation_pipeline}
\end{figure}

In GEM-STEP, we adopted a computer-vision approach for gaze estimation to observe student and teacher focus within the classroom while accommodating our logistical constraints. We encoded objects of interest (OOI) to map where participants were looking during the embodied activity, opting for a method that translated basic gaze data into more meaningful insights for our analysis. This encoding process, however, faced challenges due to the spatial nature of our learning context. To address this, we elevated our gaze analysis to a 3D perspective through room reconstruction, which provided a more accurate and physics-based approach for OOI encoding.

For the 3D room reconstruction, we utilized depth estimation to transform monocular video frames into three-dimensional space -- given that our camera was stationary. We employed ZoeDepth \cite{Bhat2023ZoeDepth}, a model trained on both indoor (NYU Depth v2) and outdoor (KITTI) datasets, for its superior depth estimation capabilities. NYU Depth v2 and KITT datasets contain 1449 and 12929 RGB and depth image pairs, respectively. Through the combination of these two datasets, ZoeDepth achieves state-of-the-art (SOTA) performance in terms of metric (absolute) depth estimation -- making it an excellent choice for reconstructing the room. This process allowed us to reconstruct each room frame-by-frame, aiding in the identification of both static (e.g., displays) and dynamic (e.g., students and teachers) OOIs. For our static OOIs, we labeled these using \href{https://github.com/ykzzyk/vision6D}{Vision6D}, a 3D annotation tool.

To minimize computational demands, we initially tracked objects in 2D before mapping them into a three-dimensional context. Utilizing face bounding boxes and cropped images produced by the MTCNN and a re-identification algorithm, we then applied the L2CS-Net \cite{Abdelrahman2022L2CSNetF}, a computer-vision model designed for gaze estimation, to calculate 3D gaze vectors for each identified face within the GEM-STEP environment. These vectors, determined by pitch and yaw measurements, were transformed into a 3D rotation matrix, denoted as $R$. By taking the $XY$ centroids of the face bounding boxes and pairing them with depth information to derive a $Z$ value, we completed the 3D translation vector $t$. The synthesis of $R$ and $t$ yielded a fully encompassing transformation matrix $RT$, encapsulating the origin and direction of a participant's gaze.

Armed with the $RT$ matrix over successive frames, we tracked the students' spatial positions and gaze directions in three dimensions over time. To account for moving objects, i.e., the students, we devised human bounding boxes anchored by the gaze's origin point, the floor's plane, and a predefined width. By employing gaze ray tracing—extending the gaze vector until it intersects with an object of interest (OOI)—and considering the room's 3D layout, we were able to encode OOIs based on where participants looked. Each frame resulted in a determined OOI for every participant, with null values assigned when faces were undetectable or gazes missed all OOIs. These OOI encodings were then compiled over 5-second intervals. Within these intervals, we adopted a mode-based pooling technique to identify and select a predominant OOI for each participant's dataset throughout the given time window.
\section{Results and Discussion}

\begin{figure}[ht]
    \centering
    \includegraphics[width=0.85\textwidth]{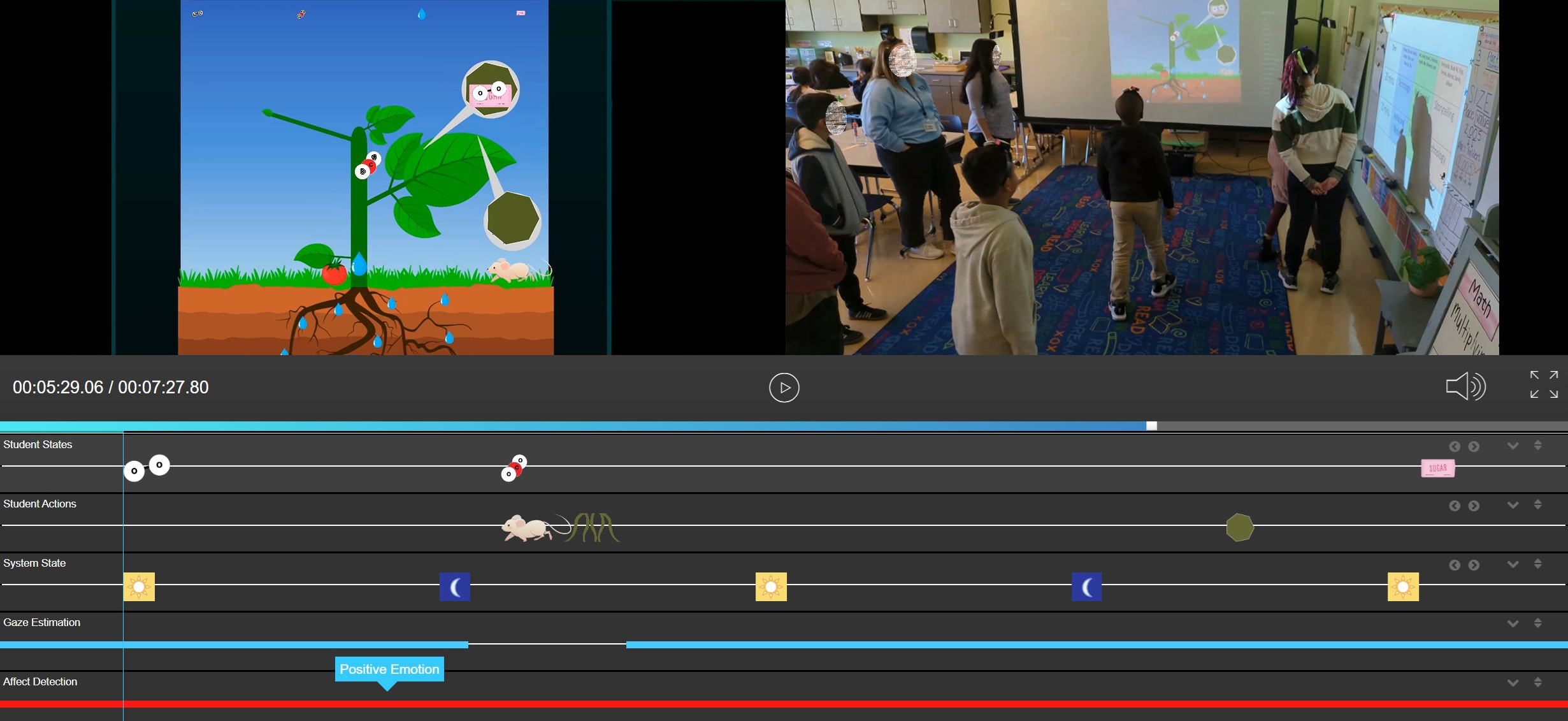}
    \caption{Visual timeline of multimodal data for IA}
    \label{fig:timeline}
\end{figure}

In our exploration of how events of IA and AI inform each other, we evaluated how our visual timeline, shown in Figure \ref{fig:timeline}, presented findings that were consistent with the interaction analysis that shaped its construction. For each video analyzed by human researchers through IA, individual student timelines were produced, allowing for new rounds of IA. The video and timeline are executed together and it is possible to navigate to specific events and zoom in/out to control the granularity of the data being shown, allowing for a more detailed or higher-level IA analysis.

The processing of system logs provided a visual representation of the molecules embodied by students in the photosynthesis process. This addressed one of the main concerns of researchers, the ability to identify components that informed students' understanding of the science model over time. By observing the molecules each student embodied and how long it took them to transition correctly through the model, researchers could pinpoint segments where students struggled and would benefit from scaffolds/feedback. Another aspect that interested researchers was how students moved in the real world and, subsequently, in the simulation. Changes in movement caused by attention shifts provide us with valuable information on how a student may be learning by interacting with the environment. This provides important links between the cognitive and sociocultural aspects of learning. Since the embodied activity required them to move around and explore the virtual environment to understand how molecules were transformed, showing visual representations of the objects of interest and when students moved towards them informed if and when students figured out the correct actions that caused transformations. Furthermore, the logs also informed system transitions of day and night time, pinpointing moments to investigate if students grasped how light affects photosynthesis. 

Currently, the temporal gaze information, which determined when and where students were looking, is plotted on the timeline, which provides cues for researchers to further investigate attention shifts. Gaze in this context was calculated and discretized to inform when students were looking at the display screen, the teacher, or one of their peers. Furthermore, such gaze behaviors coupled with conversational information (which we did not analyze in this paper) provide important information about student difficulties, e.g., if they were struggling with a specific transition, or if they received advice but chose to ignore it. In addition, our analysis and timeline representation allows us to study if gaze shifts were triggered by affect changes during the activity. Such patterns allow researchers to investigate more deeply relationships between students' affective states, their attention, how these might relate to previous actions, and how they inform future actions taken. The affect data revealed important insights. Delight was notably high among students during play, a rarity in traditional classrooms. A student who grasped the concept often felt frustrated when their peers ignored their suggestions. In the collaborative part, a knowledgeable student frequently felt sad (a subdued unpleasant emotion), seemingly due to a lack of cooperation in progressing tasks, and this was accompanied by periods of boredom. However, there were cases where once actively engaged in problem-solving, many students exhibited increased positive emotions.

Temporal data presentation was important for pattern recognition. Visual comparison of timelines from three students on the first day revealed disparate initial interactions with the model. DaPaw \footnote{none of the names used are the students' true names} required approximately five minutes for the first transition, Rose two minutes, and Taylor Swift a mere 12 seconds. IA corroborated these findings, noting DaPaw's initial hesitation and suggesting Taylor Swift's rapid transition could be explained by her prior scientific understanding, which she used to direct her peers. Moreover, our ML analysis generated aggregated metrics of student learning and performance. Rose completed the photosynthesis cycle thrice, with 15 successful molecule transitions. Overall, students spent 33\% of their initial time on carbon dioxide transformations and 21\% of their initial time on water molecule transitions. Taylor Swift seamlessly navigated all molecule transitions, completing the cycle eight times, with each of the 44 successful transitions taking under 20 seconds. Conversely, DaPaw completed the cycle once, with seven successful transitions, dedicating 66\% of the time to discern the correct action for the water molecule.

Following the initial evaluation, a group of 10 researchers started weekly collaborative sessions to further refine the tool using a user-centered approach. This process emphasized its capacity to highlight relevant segments for subsequent detailed examination that would have been hard to discern otherwise. The assessment revealed that representing students' state transitions as molecules they embody within the simulation, coupled with their navigational choices to get to parts of the screen that supported the transitions facilitated comprehension by mirroring the simulation's visual content. The tool was enhanced to permit selective modality display aligned with the investigators' specific research queries to mitigate cognitive overload from excessive on-screen data. Additionally, the tool was augmented to support the exploration of cooperative student dynamics by enabling simultaneous data visualization from multiple participants. In recognition of the activity's embodied nature, a functionality to alter the video's camera perspective in conjunction with the timeline also was integrated.
\section{Conclusion and Future Work}

Our multimodal timeline to support IA marks a significant advance in examining student interactions within mixed-reality learning settings. Leveraging machine learning, the tool captures and displays data that enriches IA, suggesting its utility in advancing IA research. It facilitates a transition from purely qualitative to mixed methods analysis by integrating quantitative data on student performance and behavioral trends over time.

Moving forward, the tool can be extended to accommodate different science models, broadening its applicability across diverse educational contexts, whether in embodied or computer-based learning environments and including modalities relevant to each context. An ongoing weekly meeting of IA sessions is currently assessing the timeline's usefulness and informing the inclusion of additional modalities, such as conversations that offer a more comprehensive understanding of communication dynamics. The iterative nature of IA, coupled with the versatility of the multimodal timeline, positions it as a dynamic framework that can evolve alongside emerging research questions and technological advancements, thereby fostering continued advancements in the field of IA applied to learning environments. We also hope to investigate how the timeline can be tailored towards the teachers and how the findings can be used to assist students during the learning activities.

\begin{credits}
\subsubsection{\ackname}This work was supported by the following grants from the National Science Foundation (NSF): DRL-2112635, IIS-1908632 and IIS-1908791. The authors have no known conflicts of interest to declare. We would like to thank all of the students and teachers who participated in this work.
\end{credits}

\bibliographystyle{splncs04}
\bibliography{references}

\begin{thebibliography}{10}
\providecommand{\url}[1]{\texttt{#1}}
\providecommand{\urlprefix}{URL }
\providecommand{\doi}[1]{https://doi.org/#1}

\bibitem{Abdelrahman2022L2CSNetF}
Abdelrahman, A.A., Hempel, T., Khalifa, A., Al-Hamadi, A.: L2cs-net : Fine-grained gaze estimation in unconstrained environments. 2023 8th International Conference on Frontiers of Signal Processing (ICFSP) pp. 98--102 (2022)

\bibitem{andrade2017understanding}
Andrade, A.: Understanding student learning trajectories using multimodal learning analytics within an embodied-interaction learning environment. In: Proceedings of the seventh international learning analytics \& knowledge conference (2017)

\bibitem{ashwin2020affective}
Ashwin, T., Guddeti, R.M.R.: Affective database for e-learning and classroom environments using indian students’ faces, hand gestures and body postures. Future Generation Computer Systems  \textbf{108},  334--348 (2020)

\bibitem{Bhat2023ZoeDepth}
Bhat, S.F., Birkl, R., Wofk, D., Wonka, P., Müller, M.: Zoedepth: Zero-shot transfer by combining relative and metric depth (2023)

\bibitem{danish2022designing}
Danish, J., Anton, G., Mathayas, N., Jen, T., Vickery, M., Lee, S., Tu, X., Cosic, L., Zhou, M., Ayalon, E., et~al.: Designing for shifting learning activities. The Journal of Applied Instructional Design  \textbf{11}(4),  169--185 (2022)

\bibitem{danish2020learning}
Danish, J.A., Enyedy, N., Saleh, A., Humburg, M.: Learning in embodied activity framework: A sociocultural framework for embodied cognition. International Journal of Computer-Supported Collaborative Learning  \textbf{15},  49--87 (2020)

\bibitem{Davalos2023chimerapy}
Davalos, E., Timalsina, U., Zhang, Y., Wu, J., Fonteles, J.H., Biswas, G.: Chimerapy: A scientific distributed streaming framework for real-time multimodal data retrieval and processing. In: 2023 IEEE International Conference on Big Data (BigData). IEEE (Dec 2023)

\bibitem{Davalos2023Identifying}
Davalos, E., Vatral, C., Cohn, C., Fonteles, J., Biswas, G., Mohammed, N., Lee, M., Levin, D.: Identifying gaze behavior evolution via temporal fully-weighted scanpath graphs. In: LAK23: 13th International Learning Analytics and Knowledge Conference. p. 476–487. Association for Computing Machinery (2023)

\bibitem{d2012dynamics}
D’Mello, S., Graesser, A.: Dynamics of affective states during complex learning. Learning and Instruction  \textbf{22}(2),  145--157 (2012)

\bibitem{enyedy2014learning}
Enyedy, N., Danish, J.: Learning physics through play and embodied reflection in a mixed reality learning environment. In: Learning technologies and the body, pp. 97--111. Routledge (2014)

\bibitem{Errea2017visual}
Errea, J., {Gestalten} (eds.): Visual journalism. Die Gestalten Verlag (Sep 2017)

\bibitem{EzZaouia2020Emodash}
Ez-zaouia, M., Tabard, A., Lavoué, E.: Emodash: A dashboard supporting retrospective awareness of emotions in online learning. International Journal of Human-Computer Studies  \textbf{139} (2020)

\bibitem{hall2015interaction}
Hall, R., Stevens, R.: Interaction analysis approaches to knowledge in use. In: Knowledge and interaction, pp. 88--124. Routledge (2015)

\bibitem{Herve2015amaliajs}
Herv{\'e}, N., Letessier, P., Derval, M., Nabi, H.: Amalia.js: An open-source metadata driven html5 multimedia player. In: Proceedings of the 23rd Annual ACM Conference on Multimedia Conference. pp. 709--712. ACM (2015)

\bibitem{gaze360_2019}
Kellnhofer, P., Recasens, A., Stent, S., Matusik, W., Torralba, A.: Gaze360: Physically unconstrained gaze estimation in the wild. In: IEEE International Conference on Computer Vision (ICCV) (October 2019)

\bibitem{kersting2021growing}
Kersting, M., Haglund, J., Steier, R.: A growing body of knowledge: On four different senses of embodiment in science education. Science \& Education  \textbf{30}(5) (2021)

\bibitem{lane2024embodied}
Lane, A., Lee, S., Enyedy, N.: Embodied resources for connective and productive disciplinary engagement [poster]. In: AERA Annual Meeting. American Educational Research Association (2024)

\bibitem{Li2020Visualization}
Li, T.H., Suzuki, H., Ohtake, Y.: {Visualization of user’s attention on objects in 3D environment using only eye tracking glasses}. Journal of Computational Design and Engineering  \textbf{7}(2),  228--237 (03 2020)

\bibitem{martinez2018physical}
Martinez-Maldonado, R., Echeverria, V., Santos, O.C., Santos, A.d., Yacef, K.: Physical learning analytics: A multimodal perspective. In: Proceedings of the 8th international conference on learning analytics and knowledge. pp. 375--379 (2018)

\bibitem{pekrun2012academic}
Pekrun, R., Stephens, E.J.: Academic emotions., p. 3–31. American Psychological Association (2012)

\bibitem{russell1980circumplex}
Russell, J.A.: A circumplex model of affect. Journal of personality and social psychology  \textbf{39}(6), ~1161 (1980)

\bibitem{savchenko2022classifying}
Savchenko, A.V., Savchenko, L.V., Makarov, I.: Classifying emotions and engagement in online learning based on a single facial expression recognition neural network. IEEE Transactions on Affective Computing  (2022)

\bibitem{Schwendimann2017Dashboards}
Schwendimann, B.A., Rodríguez-Triana, M.J., Vozniuk, A., Prieto, L.P., Boroujeni, M.S., Holzer, A., Gillet, D., Dillenbourg, P.: Perceiving learning at a glance: A systematic literature review of learning dashboard research. IEEE Transactions on Learning Technologies  \textbf{10}(1),  30--41 (2017)

\bibitem{steinberg2023making}
Steinberg, S., Zhou, M., Vickery, M., Mathayas, N., Danish, J.: Making sense of modes in collective embodied science activities. In: Proceedings of the 17th International Conference of the Learning Sciences-ICLS 2023, pp. 1218-1221. International Society of the Learning Sciences (2023)

\bibitem{tang2017multiple}
Tang, S., Andriluka, M., Andres, B., Schiele, B.: Multiple people tracking by lifted multicut and person re-identification. In: Proceedings of the IEEE conference on computer vision and pattern recognition. pp. 3539--3548 (2017)

\bibitem{ts2020automatic}
TS, A., Guddeti, R.M.R.: Automatic detection of students’ affective states in classroom environment using hybrid convolutional neural networks. Education and information technologies  \textbf{25}(2),  1387--1415 (2020)

\bibitem{Vatral2022Using}
Vatral, C., Biswas, G., Cohn, C., Davalos, E., Mohammed, N.: Using the dicot framework for integrated multimodal analysis in mixed-reality training environments. Frontiers in Artificial Intelligence  \textbf{5} (2022)

\bibitem{zhang2016joint}
Zhang, K., Zhang, Z., Li, Z., Qiao, Y.: Joint face detection and alignment using multitask cascaded convolutional networks. IEEE signal processing letters  \textbf{23}(10) (2016)

\end{thebibliography}

\end{document}